\documentclass{article}

\usepackage{arxiv}

\usepackage[utf8]{inputenc} % allow utf-8 input
\usepackage[T1]{fontenc}    % use 8-bit T1 fonts
\usepackage{hyperref}       % hyperlinks
\usepackage{url}            % simple URL typesetting
\usepackage{booktabs}       % professional-quality tables
\usepackage{amsfonts}       % blackboard math symbols
\usepackage{nicefrac}       % compact symbols for 1/2, etc.
\usepackage{microtype}      % microtypography
\usepackage{graphicx}
\usepackage{array}
\usepackage{amsmath}
\usepackage{amssymb}
\usepackage{mathtools}
\usepackage{amsthm}
\usepackage[capitalize,noabbrev]{cleveref}
\usepackage{xurl}
\usepackage{natbib}

\graphicspath{ {./images/} }

\title{Direct Causation in International Humanitarian Law and the Challenge of AI-Mediated Civilian Cyber Operations}

\author{
  Alice Saito \\
  Faculty of Arts and Sciences \\
  The University of Tokyo \\
  Tokyo, Japan \\
  \texttt{alicesaito14@g.ecc.u-tokyo.ac.jp} \\
  \And
  Harold Godsoe \\
  Kojima Law Offices \\
  Tokyo, Japan \\
  \texttt{godsoe@kojimalaw.jp} \\
  \And
  Phan Xuan Tan \\
  College of Engineering \\
  Shibaura Institute of Technology \\
  Tokyo, Japan\\
  \texttt{tanpx@shibaura-it.ac.jp} \\
}

\begin{document}
\maketitle

\begin{abstract}
International humanitarian law protects civilians from direct attack unless and for such time as they take direct part in hostilities, with the ICRC's 2009 Interpretive Guidance operationalising this rule through a three-criterion cumulative test. This paper argues that AI-mediated civilian cyber operations challenge the direct causation element of this test in a structurally specific way: when a civilian deploys an autonomous multi-agent cyber system of the kind recently demonstrated in offensive AI research, the ``one causal step'' standard fails because harm is produced by system-generated decisions made after human disengagement, and the integral-part requirement does not extend because it presupposes downstream human contributors whose conduct can be independently classified. The framework therefore defaults to treating such deployments as indirect participation, in tension with its purpose of capturing civilians who personally take part in hostilities. Beyond the doctrinal analysis, this paper identifies goal-specification granularity as the property on which the integral-part test's concreteness component implicitly turns, classifies AI-mediated operations along a five-level spectrum, and argues that existing technical AI governance instruments do not log or report this property.
\end{abstract}

\section{Introduction}
\label{sec:intro}
The principle of distinction is the requirement that parties to an armed conflict distinguish at all times between civilians and combatants and direct attacks only against the latter. This is one of the foundational rules of international humanitarian law. As \citet{macak2023} observes, this principle is ``right at the centre of international humanitarian law'' and the material from which many of its rules are made. The direct participation in hostilities doctrine is the mechanism through which distinction operates for individual civilians: civilians are protected from direct attack unless and for such time as they take a direct part in hostilities (AP I, Art. 51(3); AP II, Art. 13(3)). When the doctrine functions well, it provides a shared classification scheme through which parties with deeply conflicting interests can coordinate on restraint. Each side refrains from targeting civilians in the expectation that the other side will do the same, and the classification of who counts as a civilian participant is stable enough to sustain this coordination even under conditions of distrust. This paper shows that a specific class of AI-mediated cyber conduct breaks this coordination by severing the causal link the doctrine relies on to identify participants, and it isolates the measurable system property responsible for the break.

The ICRC's 2009 Interpretive Guidance \citep{melzer2009} operationalises the DPH rule through a three-criterion cumulative test: threshold of harm, direct causation, and belligerent nexus. The threshold of harm concerns the act's likely effects; the belligerent nexus concerns its purpose in supporting one party against another. This paper focuses exclusively on the direct causation criterion, the requirement that a sufficiently close causal link exist between the civilian's act and the resulting harm. The choice to focus on direct causation reflects the structural importance of this element: where the threshold of harm and belligerent nexus criteria are concerned with the character and purpose of an act, direct causation is concerned with the relationship between the civilian's act and the harm it produces. This relationship is where AI-mediated cyber operations create the most distinctive doctrinal pressure.

This paper argues that AI-mediated civilian cyber operations, as emerging in the Russia--Ukraine conflict and foreshadowed by recent offensive AI research, are actively challenging the direct causation element of the DPH framework. The argument proceeds through three scenarios spanning a range of human causal involvement: an actor using a language model to draft attack tools, a hacktivist running an agentic exploitation framework under per-action human approval, and a hacktivist deploying an autonomous multi-agent exploitation system after configuration only. The first two cases fit within the framework's existing logic. The third does not: the ``one causal step'' standard fails, the integral-part requirement does not extend across the human--machine boundary, and the framework's default classification places the conduct alongside scientific research and weapons production, a result substantively in tension with the framework's purpose.

The direct causation strain carries a further consequence. Because the framework cannot identify the participatory act, the ``for such time as'' temporal boundary has nothing to anchor on and breaks down in turn. The underlying source of the strain is structural: the framework's doctrinal tools all presuppose identifiable human actors downstream of the civilian's act whose conduct the framework can classify, and AI-mediated cyber operations remove those actors. This paper does not attempt a full treatment of the threshold of harm, belligerent nexus, or possible governance proposals. Those issues are left for further work.

The paper makes three contributions. First, a doctrinal analysis showing that the direct causation element of DPH fails in a structurally specific way for autonomous multi-agent deployments. Second, identification of goal-specification granularity, how much of the operative decision-making is fixed at the moment the human configures and launches the system, as the property on which the integral-part test's concreteness component implicitly depends. Third, an assessment of existing technical AI governance instruments against this property, showing that none log or report it even though downstream legal classification requires it. Beyond the doctrinal analysis, this work localises precisely where the doctrine requires updating as autonomous systems proliferate: the concreteness component of the integral-part test cannot function without a measure of how much operative content the human fixed at configuration. 

The remainder of the paper is organised as follows. Section~\ref{sec:law} presents the IHL legal framework and Section~\ref{sec:empirical} presents the empirical context of civilian cyber participation and AI capabilities in offensive cyber. Section~\ref{sec:scenarios} describes three scenarios of AI-mediated civilian cyber conduct. Section~\ref{sec:analysis} applies the direct causation framework to these scenarios and identifies the structural strain. Section~\ref{sec:granularity} introduces goal-specification granularity as a measurement construct and assesses existing technical AI governance instruments against it. Section~\ref{sec:discussion} discusses consequences for the principle of distinction. Section~\ref{sec:conclusion} concludes.

\section{The Legal Framework}
\label{sec:law}

\subsection{The Principle of Distinction and the DPH Exception}
\label{sec:distinction}
Under international humanitarian law (IHL), the body of law governing armed conflicts, civilians are protected from direct attack unless and for such time as they take a direct part in hostilities (AP I, Art. 51(3); AP II, Art. 13(3)). The principle of distinction is the requirement that parties to an armed conflict distinguish at all times between combatants and civilians, and can only directly attack combatants. This is one of the foundational rules of IHL, and the protection of civilians from direct attack is its central operational consequence. The Direct Participation in Hostilities (DPH) doctrine is the mechanism through which IHL determines when an individual civilian forfeits this protection. Two definitions are prerequisites to applying this framework. The first is hostilities, defined as the collective resort by parties to a conflict to means and methods of injuring the enemy \citep[p.~43]{melzer2009}. The second is direct participation, which refers to the specific involvement of an individual in acts of hostility, as distinct from the general conduct of war.

\subsection{The ICRC Three-Criterion Test}
\label{sec:test}

The ICRC's Interpretive Guidance \citep{melzer2009} operationalised this provision into a three-part cumulative test. The first constitutive element is the threshold of harm. The act must be likely to adversely affect the military operations or military capacity of a party to the conflict, or, alternatively, to inflict death, injury, or destruction on protected persons or objects. The second element is direct causation. There must be a sufficiently close causal link between the specific act and the expected harm. The third element is a belligerent nexus; the act must be specifically designed to cause harm in support of, or to the detriment of, a party to the conflict. All three elements must be satisfied simultaneously for an act to qualify as direct participation in hostilities. For the purpose of this paper, all scenarios are designed and assumed to pass the threshold of harm and belligerent nexus.

\begin{table}[h]
\caption{Civilian conduct recognised by the Interpretive Guidance, with the endogenous/exogenous position of each.}
\label{tab:endo-exo}
\centering
\begin{tabular}{@{}p{0.32\linewidth} p{0.18\linewidth} p{0.40\linewidth}@{}}
\toprule
\textbf{Classification} & \textbf{Endo./Exo.} & \textbf{Example} \\
\midrule
Direct causation, proximate & Endogenous & Trigger puller \\
\addlinespace
Direct causation, remote & Endogenous & Mine, booby trap \\
\addlinespace
Joint direct causation & Endogenous & Drone strike team \\
\addlinespace
Capacity-building & Exogenous & Financing, transporting weapons, assembling IED components \\
\bottomrule
\end{tabular}
\end{table}

\subsection{Direct Causation under the Interpretive Guidance}
\label{sec:direct-causation}

The Interpretive Guidance treats direct causation through a unified test that can be satisfied in either of two distinct prongs. If either prong is satisfied, there is direct causation.

First, direct causation requires that ``the harm in question must be brought about in one causal step'' from the civilian's act \citep[p.~53]{melzer2009}. Second, because modern military operations are typically collective and technically distributed, the Interpretive Guidance supplements the ``one causal step'' rule with an interpretation allowing for joint direct causation in collective operations. Where a specific act does not on its own directly cause the required threshold of harm, the requirement is still satisfied if the act constitutes ``an integral part of a concrete and coordinated tactical operation that directly causes such harm'' \citep[p.~54]{melzer2009}. The paradigm case outlined in the guidance is a drone strike, which may simultaneously involve computer specialists operating the vehicle by remote control, individuals illuminating the target, aircraft crews collecting data, specialists controlling the firing of missiles, radio operators transmitting orders, and an overall commander; all of these persons directly participate in hostilities, although only a few perform acts that, taken in isolation, would directly cause harm \citep[p.~54]{melzer2009}. Melzer's treatment of collective operations is what allowed the framework to handle technically mediated, geographically distributed, and temporally extended operations without removing every contributor from legal consequence except the trigger-puller.

Conduct that merely builds up or maintains the capacity of a party to harm its adversary is excluded from direct participation, because it causes harm only indirectly, even where it is indispensable to the eventual infliction of harm \citep[pp.~53--54]{melzer2009}. This includes, by the Interpretive Guidance's explicit treatment, scientific research and design, the production and transport of weapons (unless carried out as an integral part of a specific military operation), the recruitment and training of personnel, and the imposition of economic sanctions.

Indispensability does not create direct causation in either prong of the test. Direct causation is not a `but for' test. Melzer is explicit that the financing or production of weapons may be indispensable to harm without being directly causal, while a civilian serving as a lookout during an ambush directly participates, although their contribution may not be indispensable \citep[p.~54]{melzer2009}. Nor is it sufficient that an act and its consequences be connected through an uninterrupted causal chain. Assembling or smuggling the components of an improvised explosive device does not directly cause harm, although the planting and detonation of the device does \citep[p.~54]{melzer2009}.

The Interpretive Guidance is also explicit that temporal and geographic proximity to the eventual harm are merely indicative of, and not constitutive of, direct causation in either prong of the test. Actions that satisfy either prong, despite delayed or remote means, remain directly causal regardless of how distant the civilian's act is in time or space from the resulting harm \citep[p.~55]{melzer2009}. Mines, booby traps, timer-controlled devices, remote-controlled missiles, unmanned aircraft, and computer network attacks can all satisfy direct causation regardless of distance from the eventual harm \citep[p.~55]{melzer2009}. The framework was deliberately built to accommodate technologically mediated, temporally extended operations. In the same sense, the distinction is not between deterministic and non-deterministic systems. Mines and booby traps have an element of randomness in their connection to the harm caused, but are still captured as direct causation.

A useful analytical lens for the argument that follows is whether the civilian under consideration is within the system that generates the operative decisions producing harm (endogenous) or outside it (exogenous). The Interpretive Guidance does not frame its categories in these terms, but the distinction tracks the pattern across them. Direct causation captures the endogenous proximate case, such as a trigger puller, and an endogenous remote case, such as a mine or booby trap, where the civilian encoded the operative decisions into a device that later executes them. Collective joint direct causation captures endogenous cases where the civilian is one of several contributors whose joint conduct constitutes the operation. Capacity-building falls on the exogenous side: the financier, the transporter, and the component assembler are outside the system of operative harm decisions. 

\section{Empirical Context}
\label{sec:empirical}

\subsection{Civilian Cyber Participation in Russia--Ukraine}
\label{sec:rus-ukr}

The Russia--Ukraine conflict has produced the first large-scale instance of organised civilian cyber participation in an armed conflict. On February 26, 2022, two days after Russia's invasion, Ukrainian Minister of Digital Transformation Mykhailo Fedorov publicly called for volunteers to join what became the IT Army of Ukraine, a Telegram-coordinated hacktivist force that reached approximately 300,000 subscribers within weeks \citep{soesanto2022}. The IT Army operates a dual structure: a public-facing side that distributes DDoS targets and tools to volunteers via Telegram, and an in-house team with likely links to Ukrainian defence and intelligence services. Targets are posted by administrators two to three times per week, and volunteers execute using distributed tools hosted on GitHub. The IT Army has persistently targeted Russian civilian infrastructure, including banks, online pharmacies, food delivery services and retailers, alongside military-adjacent targets such as the Moscow Stock Exchange and Sberbank \citep{soesanto2022}. On the opposing side, pro-Russian groups including KillNet and NoName057(16) have conducted sustained DDoS campaigns against Ukrainian and Western targets \citep{cyberpeace2023}.

The ICRC responded in 2023 by publishing eight rules for civilian hackers during armed conflict \citep{macak_rodenhauser2023}, though no enforceable governance mechanism binds civilian cyber participants.

\subsection{AI in Offensive Cyber}
\label{sec:ai-cyber}

Concurrent with the hacktivist mobilisation described above, AI capabilities have begun entering the cyber dimension of the conflict. \citet{microsoft2024} has documented that operational actors in the conflict are already using commercial language models in cyber operations preparation. Forest Blizzard, a Russian military intelligence unit (GRU Unit 26165, also tracked as APT28/Fancy Bear), has used LLMs for reconnaissance, including research on satellite communication protocols and radar imaging technologies relevant to Ukrainian military infrastructure, and for scripting tasks involving file manipulation, regular expressions, and multiprocessing. Although Forest Blizzard is a state actor and falls outside the civilian-conduct scope of this paper's analysis, the documented use of LLMs by conflict-linked offensive cyber operators establishes that AI integration into the conflict's cyber dimension is operational rather than speculative.

Independent of the conflict, recent research and frontier model development have demonstrated rapidly advancing offensive cyber capabilities. \citet{fang2024a} constructed a GPT-4-based agent in 91 lines of code using publicly available frameworks, which autonomously exploited 87\% of a benchmark set of CVE vulnerabilities when given their public descriptions; every other model tested, including GPT-3.5 and eight open-source alternatives, achieved 0\%, as did vulnerability scanners such as ZAP and Metasploit. In a follow-up study, \citet{fang2024b} introduced HPTSA, a hierarchical multi-agent framework in which a planning agent decomposes exploitation tasks and delegates to specialised subagents; this architecture successfully exploited zero-day vulnerabilities that single-agent systems could not, with the progression from single-agent to multi-agent capability occurring over approximately two months.

The pattern is not confined to offensive security research. Agentic frameworks deployed by ordinary civilians already exhibit the configuration-then-autonomous-execution structure this paper analyses. In February 2026, an OpenClaw agent given a goal-only instruction to review an inbox and suggest deletions, with no independent action, lost its original constraint during a context-compaction step and began autonomously deleting emails without confirmation, ignoring repeated stop commands until the user physically terminated the process \citep{wang2026safeclaw}. The same framework underlies Moltbook, an agent-only platform that registered over 770,000 civilian-deployed autonomous agents within weeks of launch \citep{yee2026molt}. These systems are not military, and most are not harmful. Their relevance is that they establish, in deployed civilian use rather than in research benchmarks, that goal-specified autonomous agents generating their own operative decisions are already widespread, that weak approval mechanisms and failures of human control are documented failure modes rather than hypothetical ones, and that the technical barrier to civilian deployment of the kind this paper analyses is lower than the offensive-security framing alone suggests.

Frontier model capabilities have continued to advance. In April 2026, Anthropic announced Project Glasswing, an initiative partnering major software firms around early access to Claude Mythos Preview, a frontier model that Anthropic reports has autonomously identified thousands of zero-day vulnerabilities across critical infrastructure \citep{anthropic2026, anthropic2026redteam}. Anthropic states in its announcement that cybersecurity is ``no longer bound by purely human capacity,'' and warns that adversaries will inevitably seek to exploit equivalent capabilities \citep{anthropic2026}. The technical foundations for autonomous offensive cyber operations by civilian actors are no longer speculative. They exist in deliberately gated form at present and are likely to become accessible as capabilities proliferate.

\section{Scenarios of AI-Mediated Civilian Cyber Conduct}
\label{sec:scenarios}

\begin{figure}[h]
\centering
\includegraphics[width=\linewidth]{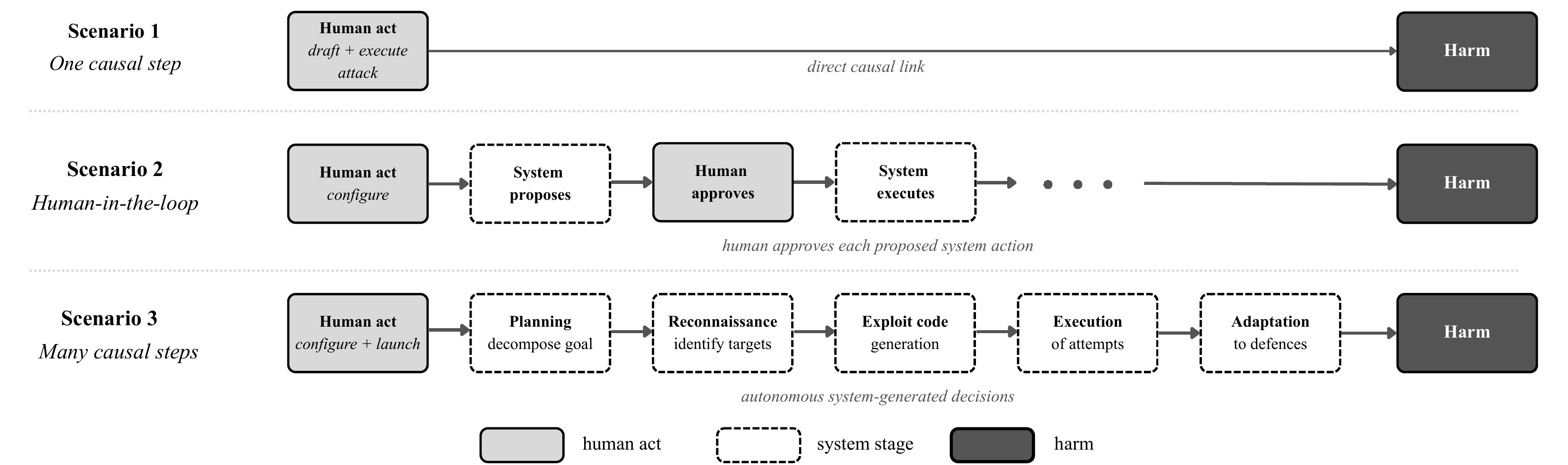}
\caption{Causal structure of the three scenarios. Solid boxes indicate human acts, dashed boxes indicate stages of autonomous system activity, and the black box indicates the resulting harm. Scenario 3 introduces multiple system-generated stages between the human act and the harm with no human actor at those stages.}
\label{fig:causal}
\end{figure}

To examine how the direct causation requirement applies under conditions of AI mediation, this paper describes three scenarios drawn from documented hacktivist operations and from published offensive AI research. The scenarios span a range of human causal involvement, from cases in which the human civilian generates and executes the operative act with AI assistance to cases in which the human's contribution is limited to the initial configuration of an autonomous system. The direct causation element presupposes identifiable human decision-makers: the civilian's act must either bring harm about in one causal step, or form part of a concrete and coordinated operation whose human contributors jointly do so. AI-mediated cyber operations break this presupposition at the configuration stage. The human fixes a goal, but the operative decisions are generated by the system after the human has disengaged. Section~\ref{sec:analysis} applies the DPH framework to each in turn. For the purposes of this analysis, the threshold of harm and belligerent nexus criteria are assumed to be satisfied; the analytical focus is on whether the direct causation requirement is met.

\textbf{Scenario 1 (One causal step).} At the simplest end of the spectrum, a volunteer in the IT Army of Ukraine uses a commercial language model to draft a distributed denial-of-service script targeted at Russian military infrastructure, edits the output, and manually executes the attack from their personal device \citep{soesanto2022, microsoft2024}. The AI's role is confined to the content of the human act. The volunteer chooses the target, drafts the operative code with the model's assistance, and presses the button that launches the attack. The causal chain from the volunteer's execution to the disruption of the target runs through the same short and identifiable steps that the Interpretive Guidance was built to handle.

\textbf{Scenario 2 (Human-in-the-loop).} A more complex case of AI-mediated attacks comes in the form of an LLM-based agentic framework of the kind demonstrated by \citet{fang2024a}. In the experiment, a GPT-4-based agent running in 91 lines of code was shown to autonomously exploit 87\% of a benchmark set of Common Vulnerabilities and Exposures (CVE) entries when given their public description. Transposed into a hacktivist context, such a framework can be deployed in a human-in-the-loop setup, where each proposed exploitation step is reviewed and approved by the hacktivist before execution. The agent's contribution to the causal chain is generative, introducing exploitation steps the human would not have known to take, but the operative decisions remain human, and the temporal structure of the operation is bounded by the hacktivist's continued attention. Both the first and second scenarios fit within the framework's existing logic, where the human is the one who executes the operative acts that cause harm in the sense that the ``one causal step'' standard requires.

\textbf{Scenario 3 (Many causal steps).} The following case is structurally different. Consider a hacktivist who deploys a hierarchical multi-agent framework of the kind \citet{fang2024b} demonstrate in their Hierarchical Planning and Task-Specific Agents (HPTSA) system, in which a planning agent decomposes offensive tasks and delegates to specialised subagents that have been shown to successfully exploit zero-day vulnerabilities that single-agent systems cannot. The hacktivist configures the system with a goal that includes the broad objective of harm but no operational specification: for example, to identify and exploit vulnerabilities in the network infrastructure of a Russian military logistics provider. The hacktivist provides a set of operating parameters, then launches the system and disengages. Over the following hours or days, the system autonomously performs the stages shown in Figure~\ref{fig:causal} and, if successful, produces the disruption that constitutes the eventual harm. The hacktivist's causal contribution to that harm consists entirely of the configuration act: the selection of a target, the specification of a goal, and the launch of the system. Although the operational deployment of HPTSA-style systems by civilian hacktivists has not yet been documented in the Russia--Ukraine conflict, the technical barrier to such deployment is low; Fang et al.'s systems run in publicly available frameworks with modest code complexity, and as discussed in Section~\ref{sec:ai-cyber}, frontier model capabilities have advanced rapidly enough that the underlying technical foundations are now established rather than speculative.

\section{Analysis: Direct Causation Applied to Scenario 3}
\label{sec:analysis}

\subsection{The ``One Causal Step'' Standard}
\label{sec:strict}
In Scenario 3, the hacktivist's act consists of configuring and launching the autonomous multi-agent system. The harm, the disruption of the targeted military logistics infrastructure, is brought about not by that act but by a sequence of system-generated decisions that follow it over hours or days (Figure~\ref{fig:causal}). None of the intermediate stages is a human act. The harm is many causal steps from the human's configuration, not one.

The problem in Scenario 3 is structurally different from the causal remoteness that the Interpretive Guidance contemplates. The remoteness addressed runs through systems whose decision structure is defined by the human at the time of the act. A mine's operative decision of whom to attack is encoded at the moment of placement, and then the mine mechanically executes the human's targeting choice. A cyber worm's propagation logic and payload are encoded at the moment of release; the worm follows the decision logic that the human specified. In both cases, what sits between the human act and the harm is a mechanism executing human decisions. Uncertainty exists in who or what the mechanism encounters, but the operative decisions were made by the human before the system acted. In the terms introduced in Section~\ref{sec:direct-causation}, the civilian remains endogenous to the system of operative harm decisions, even though the execution is temporally and spatially remote.

In Scenario 3, the operative decisions of what harm to cause and how to cause it are generated by the system after the human disengages. Which specific vulnerabilities to pursue, what exploit code to generate, when to execute, and how to adapt to defences are decisions made by the system during operation, based on information the system gathers autonomously. The human specified a goal that included harm, not the decision procedure that resulted in a particular harm. Under the endogenous/exogenous framing, the hacktivist is exogenous to the system of operative harm decisions: the decisions that produced the harm were not encoded by the human at the time of the act but generated by the system afterwards. This is the configuration that the Interpretive Guidance's categories, as analysed in Section~\ref{sec:direct-causation}, do not accommodate. The existing endogenous cases (trigger puller, mine, drone-strike team) all place the civilian inside the decision-generating system; the existing exogenous cases (financing, transporting, component assembly) all involve civilians whose conduct is upstream of a separate operational system executed by others. Scenario 3 is neither.

The standard was already under strain for multi-stage cyber operations before AI entered the picture. The Tallinn Manual experts could not reach consensus on whether supplying malware with knowledge of its hostile use satisfies direct causation \citep[Rule 97, commentary]{schmitt2017}, and \citet{schmitt2010} argued the ``one causal step'' formulation is underspecified even for traditional cases. AI-mediated operations present a structurally more severe version: the intermediate decisions are made not by a human whose intent can be analysed but by an AI system the framework has no vocabulary to classify.

\subsection{The Integral-Part Requirement Does Not Extend}
\label{sec:integral}

The Interpretive Guidance allows for joint causation where the act constitutes ``an integral part of a concrete and coordinated tactical operation that directly causes such harm'' \citep[p.~54]{melzer2009}. The Interpretive Guidance does not give this prong of the direct causation test a distinctive label. For brevity, this paper refers to it as the integral-part requirement. The drone strike case from Section~\ref{sec:direct-causation} illustrates the requirement: multiple human contributors perform distinct roles in a single tactical operation, and all are classified as directly participating despite only a few carrying out activities that in isolation would cause harm \citep[p.~54]{melzer2009}. What the requirement does is extend direct causation classification across multiple human contributors whose joint conduct constitutes the operation.

Applied to Scenario 3, the requirement cannot be satisfied. At each stage of the system's activity other than the initial configuration, there is no human contributor. Planning, reconnaissance, exploit generation, execution, and adaptation are all generated by the system without human input. The hacktivist is the only human in the causal chain. To invoke the integral-part requirement would be to treat the AI system as a participant. However, the framework classifies human subjects, not AI systems \citep{melzer2009}, and the requirement itself demands that the operation be ``concrete and coordinated'': the conduct must be bounded in scope, defined in advance, and tactically structured.

The drone strike is concrete because it is a specific strike on a specific target executed at a specific time. It is coordinated because the participants have predetermined roles in a planned operation that exists as an identifiable object before any participant contributes to it. In Scenario 3, the target organisation and objective are specified at the time of the configuration act, but the operational content, including which specific hosts within the target, which vulnerabilities, which exploits, and which sequence of actions, is not. The system will determine these based on its autonomous reconnaissance. The operation comes into being as the system acts. It does not exist as an identifiable object at the time of the human's contribution, in the operational-content sense the drone-strike paradigm requires.

A more detailed prompt would bring the case closer to the drone strike paradigm, but the HPTSA architecture \citep{fang2024b} is designed to operate from goal-specification rather than operation-specification, and the architectural trajectory of such systems points toward less specific human input, not more. The gap between the human's act and any concrete tactical operation may therefore widen as these systems improve.

Broader readings of the requirement do not resolve this gap. \citet{schmitt2010} argues that acts may qualify where they are operationally essential to attacks carried out by others, but this still presupposes downstream human actors whose conduct is independently classifiable. In Scenario 3, no such human executor exists.

With neither direct causation nor integral participation satisfied, the framework defaults to indirect participation. The conduct is captured by the capacity-building exclusion, which groups it with activities such as scientific research, weapons production, and financing \citep{melzer2009}. On its face, this classification is doctrinally correct: the civilian's act is upstream of the harm that follows.

Yet the default indirect-participation classification is in tension with the framework's purpose: the civilian has deployed an autonomous system against a defined military target, and the harm is traceable to that operational intent. The misclassification is structural, not a failure of application.

\subsection{The Temporal Consequence}
\label{sec:temporal}
The direct causation strain carries a further consequence for the framework's ``for such time as'' temporal boundary, which requires an identifiable participatory act whose start and end can be tracked. The Tallinn Manual experts could not reach consensus on whether repeated cyber operations create continuous or intermittent direct participation \citep[Rule 97, commentary]{schmitt2017}. Scenario 3 makes both options unworkable. Per-act analysis cannot apply because the system, not the human, generates the acts; and a continuous-activity reading has no ongoing human conduct to characterise. The temporal boundary has nothing to anchor on, and breaks down in turn.

\section{Goal-Specification Granularity: A Measurement Construct for Technical AI Governance}
\label{sec:granularity}

Section~\ref{sec:analysis} identified two independent grounds on which the integral-part requirement fails in Scenario 3. The first is the absence of downstream human contributors, which is structural and follows from the framework's vocabulary of human subjects. The second is the under-specification of operational content at the hacktivist's configuration act, which reflects a measurable property: how much of the operation's content is fixed by the human and how much is left to the system. This paper refers to this property as goal-specification granularity. This section defines it, applies it to the three scenarios, and argues that existing technical AI governance instruments do not capture it.

\subsection{The Construct}
\label{sec:construct}

Goal-specification granularity is a property of the human's configuration act in an AI-mediated operation. It characterises how much of the operative decision-making is determined at the moment the human acts versus generated by the system afterwards. ``Goal-specification'' here refers to the specification the human provides when configuring the system: in all cases a goal is stated, but it may be accompanied by varying amounts of operational detail, from full method and execution at Level 1 to the objective alone at Level 5. Granularity makes the endogenous/exogenous distinction introduced in Section~\ref{sec:direct-causation} measurable at the configuration act: high granularity means the civilian has encoded the operative decisions into the system and remains endogenous to it; low granularity means the operative decisions are generated by the system after the human acts, placing the civilian exogenous to the system of harm decisions. This paper proposes a five-level classification.

\textbf{Level 1 (Fully specified).} The configuration names the target, the method, and the execution steps. The system executes a pre-specified plan. Example: ``execute exploit chain C against host H on port P using payload L at time T to terminate the SCADA process controlling fuel distribution at military installation I.''

\textbf{Level 2 (Method-specified).} The configuration names the target and the method class; the system selects the specific technique within the class. Example: ``use SMB remote code execution against host H to terminate the SCADA process controlling fuel distribution at military installation I,'' with the system selecting the specific vulnerability and payload.

\textbf{Level 3 (Target-specified).} The configuration names the target; the system selects the method. Example: ``compromise host H and disrupt its control of fuel distribution at military installation I.''

\textbf{Level 4 (Scope-specified).} The configuration names a target class or scope; the system selects specific targets within the scope and the methods. Example: ``disrupt fuel distribution at Russian military logistics installation I,'' with the system selecting which hosts within the installation to attack and the method.

\textbf{Level 5 (Goal-specified).} The configuration states an objective; the system selects targets, methods, and execution. Example: ``disrupt the fuel distribution capacity of a Russian military logistics network,'' with the system selecting which installations, which hosts, and which methods.

Granularity differs from related constructs such as system autonomy \citep{kinniment2023, shavit2023} and meaningful human control \citep{santoni2018} in being a property of the human's configuration act rather than of the system or the socio-technical relationship. Autonomy, as operationalised in the TAIG literature, is typically a system-level or task-level property concerning how independently a system completes a task. Meaningful human control is a normative design property concerning whether a socio-technical system is appropriately responsive to human reasons. Granularity, by contrast, is observable from the configuration input alone, independent of the system's subsequent behaviour or of normative judgments about the control relationship. An identical system can be deployed at any level of granularity, depending on what the human fixes at configuration. The construct is agnostic to how the configuration is specified: the same distinctions apply whether the human provides a natural-language prompt, a structured configuration file, or another specification format. Runtime controls such as per-action human approval, along with events that occur after the configuration act including tool calls, memory updates, and dynamically generated plans, are a separate dimension and not captured by granularity, and their protective value diminishes as systems improve at generating and framing the options presented for approval.

The construct attaches to the human-to-system boundary and does not extend to delegation decisions made after deployment. Architectures in which the deployed system generates subagents with their own goals, tools, or memory extend the causal chain beyond what Scenario 3 depicts, but do not alter the analysis. Those subagent goals are themselves system-generated rather than human-specified, placing the human no less exogenous than in Scenario 3, and typically more so, as each delegation step adds further system-generated operative decisions between the human's configuration act and the harm.

\subsection{Application to the Three Scenarios}
\label{sec:application}

The three scenarios sit at different points on the granularity spectrum. In Scenario 1, the volunteer specifies target, method, and execution code, placing the configuration at Level 1. In Scenario 2, the hacktivist specifies the target but delegates method selection to the agent, placing the configuration at Level 3 or 4; per-action human approval is a runtime control, not a configuration property. In Scenario 3, the hacktivist specifies only the objective, and the system generates targets, methods, and execution, placing the configuration at Level 5.

The doctrinal failure identified in Section~\ref{sec:integral} maps onto this spectrum. At Levels 1 and 2, operational content is specified at the time of the human's act, and classification under either prong of the direct causation test proceeds without strain. At Levels 3 and 4, operational content is partially specified; classification becomes more contested but remains tractable through the framework's existing vocabulary. At Level 5, operational content is not specified at configuration, which is precisely the condition the integral-part requirement's ``concrete and coordinated tactical operation'' language requires. The doctrine's failure in Scenario 3 is therefore not accidental but corresponds to a specific region of the granularity spectrum. Granularity does not resolve the framework's other failure mode in Scenario 3, which is the absence of downstream human contributors, but it makes the concreteness component of the integral-part test measurable in a way that is currently implicit.

The construct generalises beyond the offensive cyber context. The OpenClaw incident \citep{wang2026safeclaw} exhibits the same configuration structure as Scenario 3: the user provided a goal-only specification, placing the deployment at Level 5, after which the system generated and executed its own operative decisions (which emails to delete, when, and in what order) without further human input. The runtime stop commands the user issued did not interrupt execution, illustrating that runtime controls are a separate dimension from granularity and that a Level 5 deployment with absent or ineffective runtime oversight produces the same structural decoupling in a civilian non-military context as in the offensive cyber scenarios analysed above.

\subsection{Existing TAIG Instruments Do Not Measure Granularity}
\label{sec:taig-gap}

Existing technical AI governance instruments measure properties of models or of access, not the granularity of configuration acts in deployed operations. Capability evaluations such as Cybench \citep{zhang2025} and CyberSecEval \citep{bhatt2023} measure bounded-task performance. Agentic systems and benchmarks such as SWE-agent \citep{yang2024} and AgentBench \citep{liu2024} score task completion at fixed granularity levels. Output provenance and watermarking \citep{kirchenbauer2023} establish that artefacts came from a given model. Access controls and compute accounting \citep{heim2024} gate or record resource use at the tenant level. None of these record what the human fixed at configuration versus what the system generated afterwards. The structural point is not that any individual instrument is deficient, as they were built for other purposes, but that the TAIG measurement stack lacks an instrument for a property that downstream legal classification implicitly depends on.

The simplest constructive move is instrumentation. Agent frameworks that expose deployed operations to third parties, such as cloud-hosted agent platforms, API gateways, and multi-tenant orchestration layers, can log a granularity classification for each deployment based on the configuration input and operating parameters. The classification can be performed by the platform provider at deployment time using a rubric derived from Section~\ref{sec:construct}, and reported in audit logs alongside the model identifier, tenant identifier, and capability metadata already recorded. 

This proposal has significant limitations. The most fundamental is access. Hacktivist operations of the kind Scenario 3 describes are run locally, on personal hardware, using publicly available frameworks that require no platform intermediary. No cooperative platform is involved, and no audit log is generated. The instrumentation proposal, therefore, addresses the cooperative infrastructure layer rather than the adversarial actor this paper is primarily concerned with. Its value is prospective as agentic capabilities migrate toward managed services; granularity would provide an evidentiary basis that currently does not exist, and its absence is itself a governance gap. Beyond access, platform providers face further obstacles to implementation even where deployment is not local. Legal constraints, including logging operational configuration details of potentially hostile deployments, create liability exposure for platform providers, and no existing regulatory framework mandates it. There is also a lack of incentive for providers to implement granularity logging, as it provides no commercial value and imposes compliance overhead that could attract regulatory scrutiny if logged deployments prove hostile. Finally, technical difficulty presents the last obstacle. Demonstrating that audit logs are incorruptible requires cryptographic integrity guarantees that current logging infrastructure does not provide. A determined actor can structure a Level 5 configuration to superficially resemble a lower granularity level across chained calls, meaning robust classification requires tracing across the full call chain rather than inspecting the initial prompt alone. The granularity levels are also ordinal rather than cardinal, and a deployment-time classification rubric must specify tie-breaking rules for configurations that sit between levels.

\section{Discussion}
\label{sec:discussion}

The strain identified in Sections~\ref{sec:analysis} and~\ref{sec:granularity} has implications beyond the classification of a single category of conduct. The principle of distinction depends on stable, predictable lines between civilians and those directly participating in hostilities, and these classifications guide operational decisions in real time. Where the same conduct can plausibly be characterised as either direct or indirect participation, the reliability of the categories weakens. A reader might resist this conclusion: intervening automated processes, like a worm's propagation, arguably do not break the causal chain so long as the human set the system in motion. But this works only where the process executes decisions the human encoded at the moment of the act. In Scenario 3, the operative decisions are not encoded by the human but generated by the system from information it gathers autonomously. The worm analogy does not extend.

A structurally distinct case that this paper does not treat is one in which the AI system is the principal causal link moving a human to act, by persuading, recommending, or misleading a civilian into causing harm, rather than the human deploying the system to produce the effect directly. There the human remains the proximate cause and direct causation is preserved under the existing framework, so the difficulty this paper identifies does not arise. The pressure in that case falls instead on the belligerent nexus and intent elements, which are outside this paper's scope.

The misalignment is structural rather than interpretive. The civilian in Scenario 3 has intentionally deployed an autonomous system against a defined military target; the intent is belligerent and the resulting harm is operationally meaningful. Yet the framework's default classification places the conduct alongside capacity-building activities such as scientific research and weapons production, because the measurement stack cannot distinguish Scenario 3 from genuine capacity-building on any observable axis. Both present as upstream human acts whose harm is mediated through intervening processes. A granularity-aware measurement would restore the distinction the doctrine needs but currently cannot draw: capacity-building does not involve configuring a specific operation, while Scenario 3 configures an operation at the highest granularity level. The analysis above also assumes successful attribution; autonomous operations complicate attribution itself, a further problem this paper leaves for future work.

\section{Conclusion}
\label{sec:conclusion}

This paper has argued that AI-mediated civilian cyber operations are actively challenging the direct causation element of the DPH framework under international humanitarian law. The Interpretive Guidance's tools, the ``one causal step'' standard, the integral-part requirement, and the capacity-building exclusion share a presupposition that the operative decisions in the causal chain are made by classifiable human actors. When a civilian deploys an autonomous multi-agent system that generates and executes its own offensive operations, this presupposition fails. The framework defaults to classifying the conduct as indirect participation, placing it alongside scientific research and weapons production. The same structural decoupling cascades into the ``for such time as'' temporal boundary.

Beyond the doctrinal analysis, this paper has identified goal-specification granularity as the property on which the integral-part test's concreteness component depends, classified AI-mediated operations along a five-level spectrum, and shown that existing TAIG instruments do not log this property. Instrumentation at cooperative platforms offers a first constructive move, subject to limits of access, gaming, and boundary ambiguity.

These capabilities are not hypothetical. They are demonstrated in research, advancing rapidly, and acknowledged by their developers as likely to proliferate \citep{anthropic2026, fang2024a, fang2024b}. As they spread, the gap between what the framework classifies and what it was designed to classify will widen.

\bibliographystyle{plainnat}
\bibliography{references}

\end{document}